\pdfoutput=1
\documentclass[nohyperref]{article}

\usepackage{microtype}
\usepackage{graphicx}
\usepackage{enumitem}
\usepackage{booktabs} %
\usepackage{makecell}
\usepackage{multirow}
\usepackage{multicol}
\usepackage{hhline}
\usepackage{url}
\usepackage{hyperref}

\newcommand{\changed}[1]{#1}

\def\xvar{{\sf {\bf x}}}
\def\yvar{{\sf {\bf y}}}

\newcommand{\tent}{Tent}
\newcommand{\cpl}{Conjugate PL}
\newcommand{\memo}{MEMO}
\newcommand{\rpl}{Robust Pseudo-Label}

\newcommand{\healthdata}{Cardiovascular}
\newcommand{\imagenetx}{ImageNet-X}
\newcommand{\kitti}{KITTI}
\newcommand{\cityscape}{Cityscapes}
\newcommand{\physionet}{Physionet}
\newcommand{\PhraseBank}{PhraseBank}

\newcommand{\tool}{SQRL}
\newcommand{\toolfull}{Statistical Quantile Rule Learning}

\usepackage{algorithmic}
\usepackage[accepted]{icml2023}

\usepackage{amsmath}
\usepackage{amssymb}
\usepackage{mathtools}
\usepackage{amsthm}

\usepackage{caption}
\usepackage{subcaption}

\usepackage[capitalize,noabbrev]{cleveref}

\theoremstyle{plain}

\theoremstyle{definition}

\theoremstyle{remark}

\icmltitlerunning{Do Machine Learning Models Learn Statistical Rules Inferred from Data?}

\begin{document}

\setlength{\abovedisplayskip}{8pt}
\setlength{\belowdisplayskip}{8pt}
\setlength{\abovedisplayshortskip}{8pt}
\setlength{\belowdisplayshortskip}{8pt}

\twocolumn[
\icmltitle{Do Machine Learning Models Learn Statistical Rules Inferred from Data?}

\begin{icmlauthorlist}
\icmlauthor{Aaditya Naik}{upenn}
\icmlauthor{Yinjun Wu}{upenn}
\icmlauthor{Mayur Naik}{upenn}
\icmlauthor{Eric Wong}{upenn}
\end{icmlauthorlist}

\icmlaffiliation{upenn}{Department of Computer and Information Science, University of Pennsylvania, PA, USA}

\icmlcorrespondingauthor{Aaditya Naik}{asnaik@seas.upenn.edu}

\vskip 0.3in
]

\printAffiliationsAndNotice{}  %

\begin{abstract}
Machine learning models can make critical errors that are easily hidden within vast amounts of data.
Such errors often run counter to rules based on human intuition.
However, rules based on human knowledge are challenging to scale or to even formalize.
We thereby seek to infer statistical rules from the data and quantify the extent to which a model has learned them.
We propose a framework SQRL
that integrates logic-based methods with statistical inference to derive these rules from a model's training data without supervision.
We further show how to adapt models at test time to reduce rule violations and produce more coherent predictions.
SQRL generates up to 300K rules over datasets from vision, tabular, and language settings.
We uncover up to 158K violations of those rules by state-of-the-art models for classification, object detection, and data imputation.
Test-time adaptation reduces these violations by up to 68.7\% with relative performance improvement up to 32\%.
SQRL is available at \url{https://github.com/DebugML/sqrl}.

\end{abstract}

\vspace{-0.15in}
\section{Introduction}
\label{sec:intro}

Machine learning models can make a variety of errors due to factors such as noisy data, poor model generalizability, and domain shift.
\changed{Understanding} a model’s performance with respect to such errors typically involves the use of quantitative metrics such as accuracy or F1-score.

While these metrics assess a model’s performance in an overall sense, they lack the ability to distinguish fundamental errors from mundane ones.
Indeed, a model with lower accuracy could even make more coherent predictions than a model with higher accuracy.
Making datasets and models even bigger to improve upon these metrics
further exacerbates this problem.
The net result is that the most critical errors are easily hidden within vast amounts of data.

\begin{figure}
    \centering
    \includegraphics[width=\linewidth]{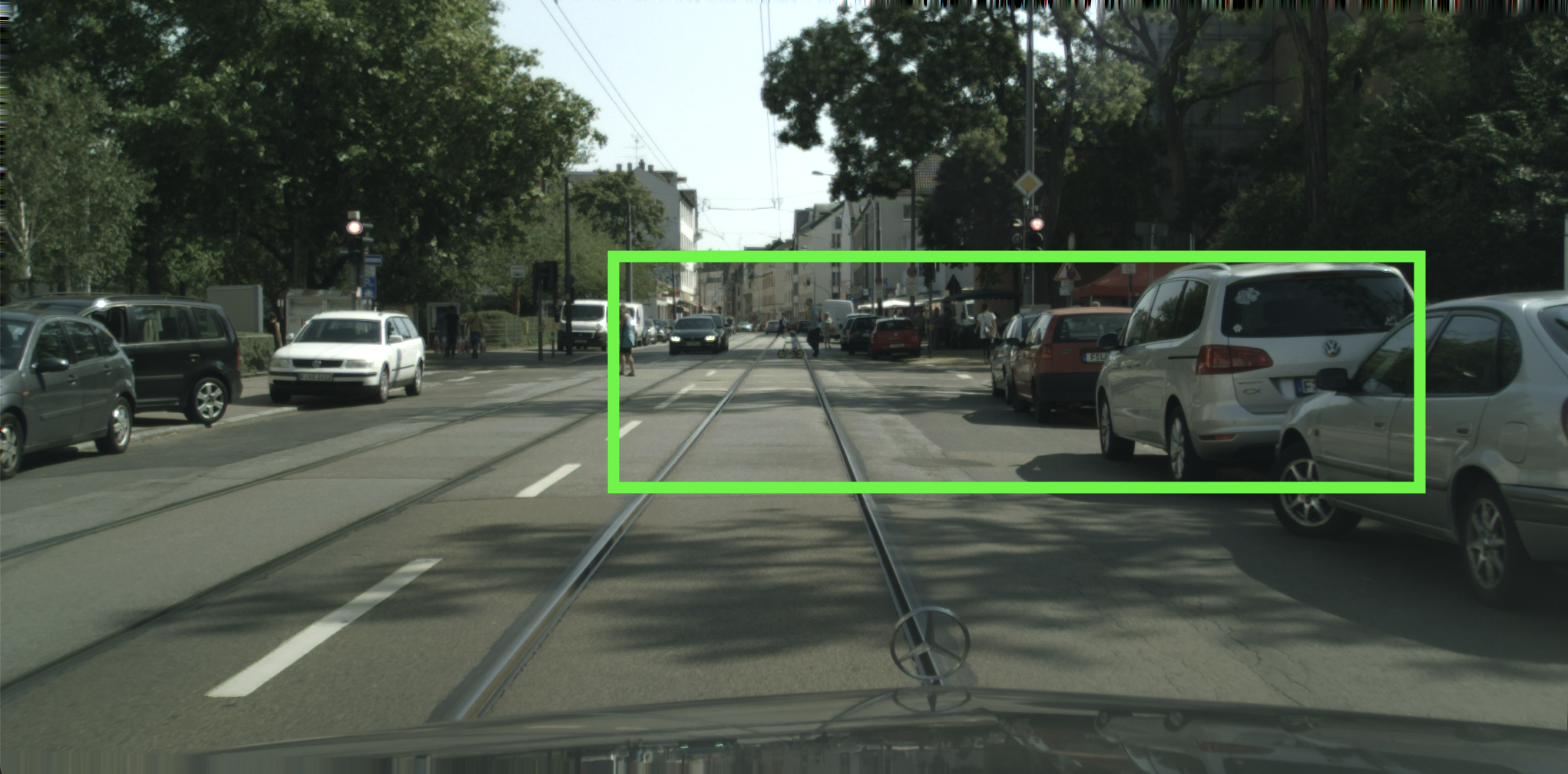}
    \vspace{-0.2in}
    \caption{The bounding box predicted as a car illustrates a \changed{basic} error by EfficientPS, a top performing model for panoptic segmentation in the \cityscape\ challenge. 
    }
    \label{fig:motivational_figure}
    \vspace{-0.15in}
\end{figure}

As an example, consider Figure~\ref{fig:motivational_figure}, where the green bounding box spanning across the frame is predicted to be a car by EfficientPS \cite{mohan2021efficientps}, a top-performing
 model for panoptic segmentation on the \cityscape\ dataset \cite{cordts2016cityscapes}.
This prediction is an instance of a fundamental error since it obviously defies reasonable notions of \changed{the shapes of} cars.
Such an error may even cause the autonomous vehicle's controller to halt the vehicle abruptly with harmful consequences.

These errors often run counter to rules based on human intuition (e.g., intuition about the width of a typical car with respect to its height).
We can therefore cast the problem of estimating such errors in terms of finding rule violations.
Such rules can not only help in estimating these errors but also in preventing them by improving the model's predictions with respect to the rules.

A central challenge concerns how to identify such rules at scale.
On one hand, the rules must be statistically valid with a well-defined meaning to avoid reporting uninteresting or false violations (i.e., false positives). 
On the other hand, the rules must go beyond basic concepts and express complex phenomena in the data to avoid missing interesting violations (i.e., false negatives).
Even human experts may fail to come up with or agree upon such a set of rules, including identifying relevant statistics, their bounds, and logical predicates over them.

Our key insight is to synthesize a set of statistical quantile rules by statistically \changed{deriving} rules that the training data conforms with.
Our approach not only automates the effort to specify a large set of rules but also to determine the goodness of a rule.
We have also implemented a general framework, \toolfull{} (\tool), which integrates logic-based methods with statistical inference to synthesize such rules from a given dataset without supervision.
We also propose to adapt models at test time to reduce violations of the synthesized rules.

We evaluate our approach by applying it to datasets and models from five different domains: tabular classification on a cardiovascular disease dataset \citep{cardio},
image classification on ImageNet \citep{deng2009imagenet}, object detection on the \cityscape\ \cite{cordts2016cityscapes} and KITTI \cite{geiger2013vision} datasets, time-series data imputation over the Physionet dataset~\citep{silva2012predicting}, and sentiment analysis over the Financial PhraseBank dataset~\citep{malo2014good}.
Our framework generates between 35 to around 300K rules over these datasets and finds between 578 to around 158K violations of those rules on the model predictions.
To correct those rule violations, we adapt the models at test time, which reduces the rule violations by up to 68.7\% and achieves a relative performance improvement by up to 32\%.

We summarize the contributions of our work:
\begin{enumerate}[leftmargin=*,nolistsep, nosep]
    \item We propose statistical quantile rules that estimate \changed{basic} errors for machine learning datasets and models.
    \item We develop a framework \tool{} to synthesize valid and expressive rules from data without additional supervision.
    \item We propose to improve the models by reducing violations of the rules using test-time adaptation.
    \item We evaluate our approach using the rules as metrics over models trained for five different applications.
\end{enumerate}

\section{Our Approach}
\label{sec:problem}

Using rules to characterize \changed{basic} errors in machine learning is a recurring strategy for improving the trustworthiness and fidelity of models.
For example, knowledge graphs and logical constraints have been used for integrating domain knowledge into machine learning pipelines. While these approaches have worked in small and controlled settings, they face several challenges in being applicable to large data settings. Knowledge graphs require substantial human effort to construct even for small datasets, while hard logical constraints can be too simple for noisy datasets often filled with exceptions.

We thereby seek an alternative way that is suited for modern machine learning datasets. 
To answer this question, we first identify the key desiderata of a representation of such rules:
\begin{enumerate}[leftmargin=*,nolistsep,nosep]
\item \textbf{Validity:} they must be valid for most of the data, but allow for exceptions or noise,
\item \textbf{Expressivity:} they must be capable of expressing or capturing complex phenomena and relations, and 
\item \textbf{Scalability:} they must be generatable without requiring human supervision on individual rules. 
\end{enumerate}
We illustrate these criteria using the example from Figure~\ref{fig:motivational_figure}. 
Validity requires that akin to human intuition, any rule that enables the detection of this particular mistake must be true for most other images as well, so as to not raise false alarms.
In our example, it is reasonable that the shape of the bounding box is unlikely to represent a car.
Expressivity is a competing requirement that states that the representation must be expressive enough to avoid only the most basic or vacuous rules that miss these errors.
For example, if using the shape of the car alone is inadequate, it should be possible to use a combination of attributes, such as different ranges of shapes depending on the position of the car.
Finally, scalability acknowledges that manually crafting such rules for rich datasets is infeasible, as it involves a combinatorial explosion of features to consider along with statistically valid bounds on them.

\subsection{Statistical Quantile Rules}
We propose quantile-based statistical inference as a general framework for representing and generating such rules. Specifically, let $X$ be a random variable for a data point, and let $\phi(X)$ be some statistic of $X$. Then, if
\begin{equation}
\label{eq:quantile-rule}
\mathbb P\left(a \leq \phi(X) \right) = 1-\delta,
\end{equation}
for some $\delta$ and $a$, we say that $a\leq \phi(X)$ is a $1-\delta$ quantile rule. In other words, this means that $1-\delta$ of all the data has a statistic $\phi(X)$ that falls above $a$. 

For example, suppose $X$ is a random variable representing bounding boxes of cars from the \cityscape{} challenge in Figure~\ref{fig:motivational_figure}. Possible statistics $\phi(X)$ for a car are its width, height, or aspect ratio. To obtain the quantile rule for the aspect ratio statistic, we can calculate the aspect ratio of every car in the training data, and find the $1-\delta$ quantile to get the threshold $a$.

Quantile rules satisfy the above desiderata. First, they are valid for $1-\delta$ of the data, with the $\delta$ fraction allowing for exceptions. Second, a quantile rule can be computed for \emph{any} kind of statistic, which can be arbitrarily complex. While the example demonstrates a quantile rule for basic shape properties such as height and width, one can consider more complex statistics such as lighting, textures, or even predictions from other machine learning models. Most importantly, a quantile rule does not need a human to check if a given rule is good or bad. The rule is, by construction, a valid $1-\delta$ quantile for a statistic on the training data.\footnote{One can also easily check if the quantile rule generalizes by checking the rule on a validation set.} A ``good'' quantile rule provides a tight threshold $a$, whereas a ``bad'' quantile rule results in a very conservative threshold $a$.  Although overly conservative thresholds may not be very useful (and may be vacuous), all quantile rules circumvent the need for a human to check for correctness as they are correct by construction.   

\subsection{Classes of Quantile Rules}
\label{sec:rules:classes}
The quantile rule framework can capture a wide range of knowledge by varying the statistic $\phi$. To highlight the expressivity of the framework, we discuss several general classes of quantile rules that can be viewed as variants of the original quantile rule from \eqref{eq:quantile-rule} that we use in this paper. 

{\bf Two-sided quantile rules.} A simple generalization of the quantile rule is to use a double-sided quantile to get both an upper and a lower bound for a given statistic. Specifically, if
\begin{equation}
\mathbb P\left(a \leq \phi(X) \leq b\right) = 1-\delta, 
\end{equation}
then $\phi(X) \in [a,b]$ is a $1-\delta$ quantile rule. Typically, we can compute $a$ and $b$ to be the lower and upper $\frac{1-\delta}{2}$ quantiles respectively. For example, we can compute not just a lower bound on the aspect ratio, but a complete interval of common aspect ratios. 

{\bf Mini-batch quantile rules.}
A generalization of the quantile rule is to consider statistics of not just one but a minibatch of random variables. Specifically, suppose $\mathbf{X} = (X_1, \dots, X_m)$ is a minibatch of size $m$. Then, if 
\begin{equation}
\mathbb P\left(a \leq \phi(\mathbf X)\right) = 1-\delta, 
\end{equation}
then $a\leq \phi({\mathbf X})$ is a minibatch quantile rule, where $\phi$ can be any minibatch statistic such as mean or standard deviation.\footnote{The $1-\delta$ quantile for a minibatch statistic over randomly resampled minibatches can be interpreted as a classic $1-\delta$ confidence interval.}

{\bf Logic quantile rules.} 
The statistic $\phi$ can be much  more than a mathematical formula---it can also involve discrete, logical expressions. Specifically, let $\psi$ be a Boolean logical formula, and let $\psi(\mathbf X)$ be a formula evaluated on every example of a minibatch $\mathbf X$. Then, if 
\begin{equation}
\mathbb P\left(a \leq \phi(\psi(\mathbf X))\right) = 1-\delta, 
\end{equation}
then $a \leq  \phi(\psi(\mathbf X))$ is a logic quantile rule. For example, $\psi$ could  be the logical formula $\texttt{smoke}(x) \Rightarrow \texttt{has\_cardio\_disease}(x)$, and $\phi$ could be any minibatch summary statistic such as the F1 score  of this rule. 

{\bf Neural quantile rules.}
To highlight the expressivity of the statistic, we can consider quantile rules that use statistics of another model, i.e., a neural network. Specifically, let $f$ be a neural model that extracts some features (such as an object detector or an attribute extractor). Then, if  
\begin{equation}
\mathbb P\left(a \leq \phi(f(X))\right) = 1-\delta, 
\end{equation}
then $a\leq \phi(f(X))$ is a neural quantile rule. For example, $f$ could be an object detector that extracts various objects such as cars, signs, and pedestrians, while $\phi$ can be any statistic of the resulting objects such as counts or sizes.

Lastly, we can combine concepts from the above classes of rules to produce even more expressive ones.
We illustrate different combinations in the rest of the paper.

\section{The \tool{} Framework}
\label{sec:framework}

\begin{figure*}
    \centering
    \includegraphics[width=\textwidth]{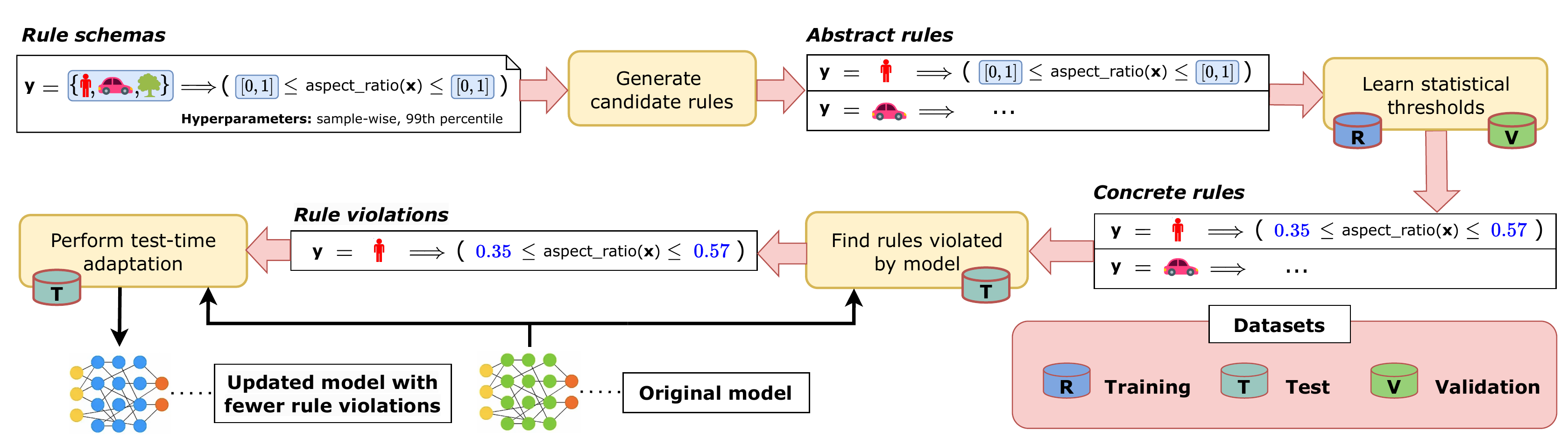}
    \vspace{-0.15in}
    \caption{Workflow of our \tool{} framework for generating rules and evaluating and adapting models for rule violations.} 
\label{fig:workflow}
\vspace{-0.1in}
\end{figure*}

The space of statistical quantile rules possible for a dataset is unbounded.
A large dataset with rich features can induce an enormous set of such rules even with a small bound on the rule size.
It is thus infeasible to rely on human supervision to manually craft and check each rule one by one.

We, therefore, develop an end-to-end framework, \toolfull{} (\tool{}), to generate rules at scale.
The workflow of \tool{} is depicted in Figure \ref{fig:workflow}.
We first describe how it generates rules from data and then present how the generated rules can be used to evaluate and improve a model's violations of those rules.

\subsection{Generating Statistical Quantile Rules}\label{sec: rule_learning}

Our formalism of statistical quantile rules is designed to be applicable to a rich variety of machine learning datasets and models.
As such, it is too general for any specific scenario, and we desire a mechanism to enable a domain expert to guide the generation of rules.
For instance, the choice of generating logic versus neural quantile rules depends on whether to use a statistic over a logical formula or a neural model's output.
Likewise, the choice of sample-based versus mini-batch quantile rules depends on whether a statistic of individual samples or mini-batches is more appropriate.

The power of statistical quantile rules can be attributed to two distinct sources of unboundedness: the structure of the rules and
the statistical bounds.
Consider the following concrete rule over a sample $(\xvar, \yvar)$:
\[
\yvar = {\sf person} \Rightarrow (0.35 \leq {\sf aspect\_ratio}(\xvar) \leq 0.57),
\]
where $\xvar$ is an image with a single bounding box derived by a neural model, ${\sf aspect\_ratio}$ is the aspect ratio of the box, and $\yvar$ is the object label.
This rule illustrates a conditional extension of a two-sided neural quantile rule.
We use this form of the rule when the learned bounds apply to a subset of samples that satisfy a condition, e.g., $\yvar = {\sf person}$.
 
This rule is one of an unbounded number of possible rules in terms of the structure (e.g., choice of neural models, features, and logic predicates) and the statistical bounds (e.g., the interval of the aspect ratio).

\tool{} allows the user to succinctly specify a {\em rule schema} $S$ to guide rule generation.
The rule schema serves as a prior over the generated rules, analogous to
meta-rules in Inductive Logic Programming \cite{Metagol}, mode declarations in Answer Set Programming \cite{law2020ilasp}, and rule templates in logic program synthesis \cite{popl20}.
Then, the problem we wish to solve can be stated more formally as follows:

Given a rule schema $S$ and a dataset $D$, output a set of concrete rules $R$ that is:
\begin{enumerate}[nolistsep, nosep]
\item {\em consistent} with $S$: no generated rule can include constructs not specified within $S$,
\item {\em exhaustive} over $S$: every single rule consistent with $S$ must be generated, and
\item {\em valid} with respect to $D$: each generated rule must hold over $1 - \delta$ of $D$.
\end{enumerate}

\tool{} solves this problem in two phases, using a combination of logic-based methods and statistical inference.
In the first phase, it uses logic-based methods to enumerate a set of {\em abstract rules} with respect to the user-provided rule schema $S$.
In the second phase, it uses statistical inference to compute bounds for each abstract rule to yield a concrete rule that is valid according to $D$.

The input rule schema $S$ is a set of rule templates of the form $T[\Theta]$ where $T$ specifies the structure of the rule, including any hyperparameters (e.g., mini-batch size and the percentile threshold), and $\Theta$ specifies placeholders for the statistical bounds.
Rules of different classes defined in Section \ref{sec:rules:classes} require different forms of schemas.
Our running example uses the following schema for two-sided quantile rules:
\vspace{-0.03in}
 \begin{align*}
    Y & \in \{ \sf{car},\ \sf{person},\ \sf{rider},\ \ldots \}, \ \phi \in \{ \sf{aspect\_ratio},\ \ldots \}, \\
    \yvar &= Y \Rightarrow \theta_{\sf lb} \leq \phi(\xvar) \leq \theta_{\sf ub}, \ 1-\delta = 0.98,
\end{align*}
\vspace{-0.02in}
where $Y$ is the set of object labels, $\phi$ is the set of statistics, and $\Theta = [\theta_{\sf lb}, \theta_{\sf ub}]$ are placeholders for the lower and upper bounds on the statistic for a given label for the percentile threshold specified by $1-\delta$.

The above schema results in a set of several abstract rules:
\vspace{-0.03in}
\begin{align*}
    \yvar = \sf{car} & \Rightarrow \theta_{\sf lb, car} \leq {\sf aspect\_ratio}(\xvar) \leq \theta_{\sf ub, car}, \\
    \yvar = \sf{person} & \Rightarrow \theta_{\sf lb, person} \leq {\sf aspect\_ratio}(\xvar) \leq \theta_{\sf ub, person}, \\
    \yvar = \sf{rider} & \Rightarrow \theta_{\sf lb, rider} \leq {\sf aspect\_ratio}(\xvar) \leq \theta_{\sf ub, rider}.
\end{align*}
The algorithm for generating these abstract rules takes $O(mn^k)$ time, where $m$ is the number of labels in $Y$, $n$ is the number of statistics in $\phi$, and $k$ is the largest number of statistics to be used in a single rule.

Concrete rules are then instantiated from each abstract rule by learning the 98th percentile bounds over the given dataset. 
An example of such a rule generated over the KITTI dataset \cite{geiger2013vision} is:
\begin{equation*}
\label{eq:rule_example}
   \yvar = \sf{car} \Rightarrow 0.07 \leq {\sf aspect\_ratio}(\xvar) \leq 2.77.
\end{equation*}
The method to obtain such bounds for each abstract rule is outlined in Algorithm \ref{alg:compute_stat}.
It calculates each abstract rule's statistics over multiple randomly drawn samples (or mini-batches) from the training set, over which the statistical bounds are obtained.
Since the bounds may not hold over unseen datasets, we only keep the rules for which the statistical bounds are consistent between the training set and a held-out validation set according to the Jaccard index, which is formulated in Appendix \ref{sec: app_tech}.
For each rule, consider $n$ to be the size of each minibatch. To derive the quantile bounds for that minibatch, we must first sort the statistics, which takes $O(n \text{log}(n))$ time.
We learn the bounds themselves using the \texttt{numpy.percentile} function. While its complexity is not stated, assuming it is $O(n)$, deriving the bounds for each minibatch takes $O(n \text{log}(n))$ time.

\begin{algorithm}
\begin{small}
\caption{Computing statistics for abstract rules}
\label{alg:compute_stat}
\textbf{Input}: {Training set $R$, validation set $V$, a set of abstract rules $\mathcal{R}_{\text{abs}}$, mini-batch size $B$, a quantile threshold $1-\delta$, a validation threshold $\epsilon$}\\
\textbf{Output}: {A set of candidate rules: $\mathcal{R}_{\text{cand}}$}
\begin{algorithmic}[1]

    \STATE Randomly sample $N_1$ mini-batches of size $B$ from $R$

    \STATE Randomly sample $N_2$ mini-batches of size $B$ from $V$

    \STATE Initialize the set of output rules as empty: $R_{\text{cand}}=\{\}$
    
    \FOR{each rule $r \in R_{\text{abs}}$}
        \STATE collect statistics over each of the above $N_1$ training mini-batches for $r$;
        \STATE compute $1-\delta$ percentile bounds, $\phi_{\text{train}}$, over those samples;
        \STATE collect statistics over each of the above $N_2$ validation mini-batches for $r$;
        \STATE compute $1-\delta$ percentile bounds, $\phi_{\text{valid}}$, over those samples;

        \IF{$|\text{Jaccard}(\phi_{\text{train}}, \phi_{\text{valid}})| > 1 - \epsilon$}
            \STATE add $r$ to $\mathcal{R}_{\text{cand}}$
        \ENDIF
    \ENDFOR
\end{algorithmic}
\end{small}
\end{algorithm}

\subsection{Using the Statistical Quantile Rules}
\label{sec:uses}
There are many potential applications of the rules generated by \tool{}.
In this paper, we focus on two important uses: evaluating and improving models with respect to statistical quantile rule violations.

To evaluate rule violations, we establish a metric
that calculates the number of violations of each rule in the model's predictions on a test set.
A rule is violated if the value of $\phi(X)$ lies outside the learned bounds, where $X$ is either a single sample or a minibatch depending on the rule.

We can also attempt to improve the number of violations of a given model on the statistical quantile rules by incorporating an auxiliary semantic loss that captures rule violations. However, it can be impractical to re-train or fine-tune a large model using such a loss, since our framework typically generates a large set of rules.
We therefore instead adapt the model's performance at test time.

Consider a statistical quantile rule $r$ of the form $\theta_{\sf lb} \leq \phi(X)$.
We define a loss depending on model parameters $W$ for its output $X$ over a sample (or a mini-batch of samples) as so:
\begin{align}
    \label{eq:loss_general}
    l_W(r, X) = \begin{cases}
    0 & \text{if}\ X\ \text{satisfies}\ r \\
    \min \{ \theta_{\sf lb} - \phi(X), 1 \} & \text{otherwise} \\
    \end{cases}
\end{align}   
Essentially, the loss is a zero or non-zero value assigned to those samples satisfying and violating $r$ respectively.
Intuitively, the non-zero loss is the distance between $a$ and the result $\phi(X)$ of evaluating the rule $r$ over the sample $X$.
To avoid potentially large loss values, the loss is clipped at~1.
The loss functions for other kinds of rules are similarly defined in Appendix \ref{sec: two-side-loss}.
Given a set of concrete rules $\mathcal{R}$, we average Equation \eqref{eq:loss_general} for all of them as the objective function for test-time adaptation:
\begin{align}\label{eq: loss}
    \mathcal{L}_W(\mathcal{R}, \mathbf X) = \frac{1}{|\mathcal{R}|}\sum\nolimits_{r \in \mathcal{R}} l_W(r,\mathbf X).
\end{align}    
The above loss may not be differentiable for some statistics $\phi$, such as the F1 score.
In such cases, we use their conjugates, such as the conjugate F1 score \cite{benedict2021sigmoidf1}.

Algorithm \ref{alg:tta_by_rule} sketches the overall test-time adaptation algorithm for reducing rule violations. It evaluates Equation \eqref{eq: loss} in each iteration and performs back-propagation on all parameters from the batch-normalization layers.
The process is repeated for $N$ iterations in total.

\begin{algorithm}
\begin{small}
\caption{Rule-based test-time adaptation algorithm}
\label{alg:tta_by_rule}
\textbf{Input}: {Test dataset $T$, a set of statistical quantile rules $\mathcal{R}$, and a pretrained model $M$, number of iterations $N$}
\begin{algorithmic}[1]
    \STATE \textbf{Initialization} collect the parameters from all the batch normalization layers of $M$ as a set $W$ and only fine-tune $W$, keeping all the other parameters in $M$ fixed
    \WHILE{number of iterations is smaller than $N$}

    \STATE Randomly sample a mini-batch $\mathbf{X}$ from $T$
    
    \STATE Evaluate loss $\mathcal{L}_W(\mathcal{R}, \mathbf X)$ using Equation \eqref{eq: loss}\\

    \IF{$\mathcal{L}_W(\mathcal{R}, \mathbf X) > 0$}

        \STATE Perform back-propagation on $M$ and update $W$
    
    \ENDIF
    \ENDWHILE
\end{algorithmic}
\end{small}
\end{algorithm}

\section{Evaluation}

\begin{figure*}[]
     \centering
     \begin{subfigure}[b]{0.32\textwidth}
         \centering
         \includegraphics[width=\textwidth, height=1.52\textwidth]{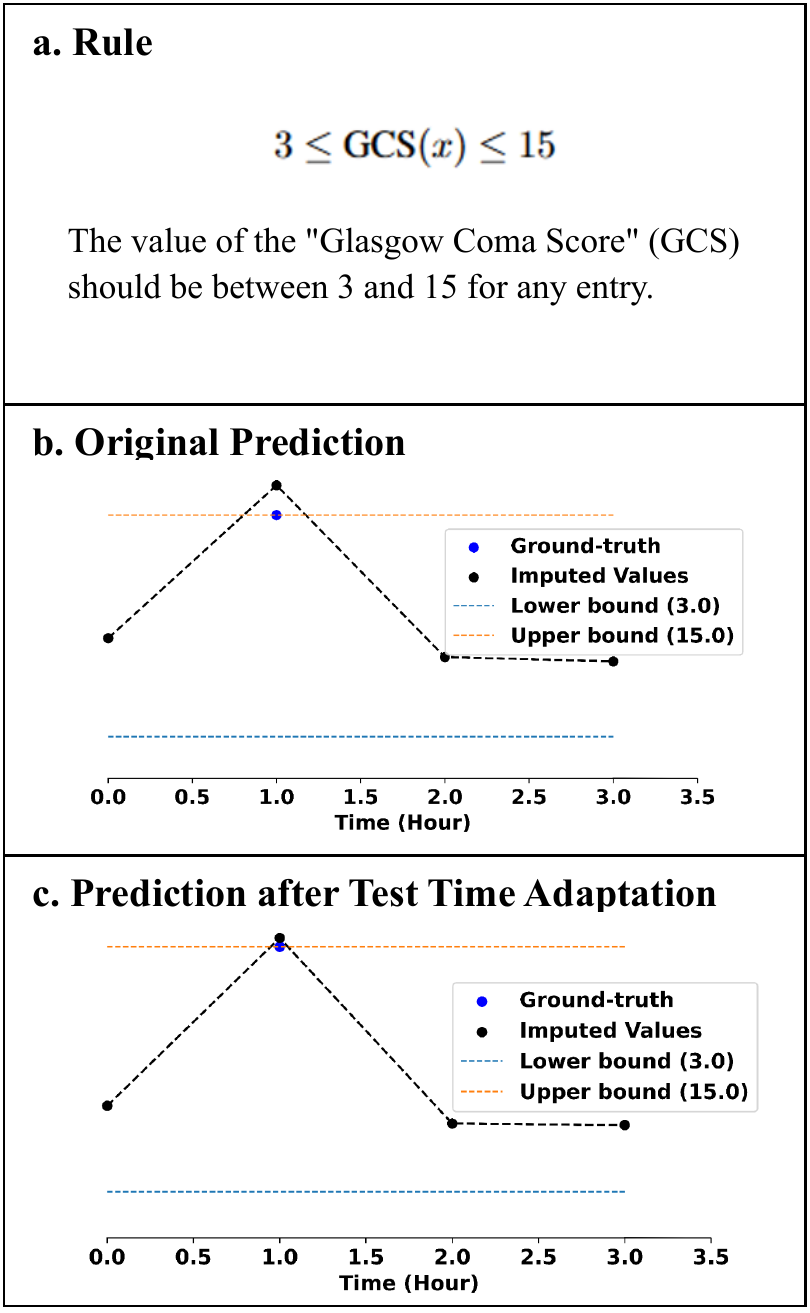}
         \caption{Time series imputation.}
         \label{fig:qual:time_series}
     \end{subfigure}
     \hfill
     \begin{subfigure}[b]{0.32\textwidth}
         \centering
         \includegraphics[width=\textwidth, height=1.52\textwidth]{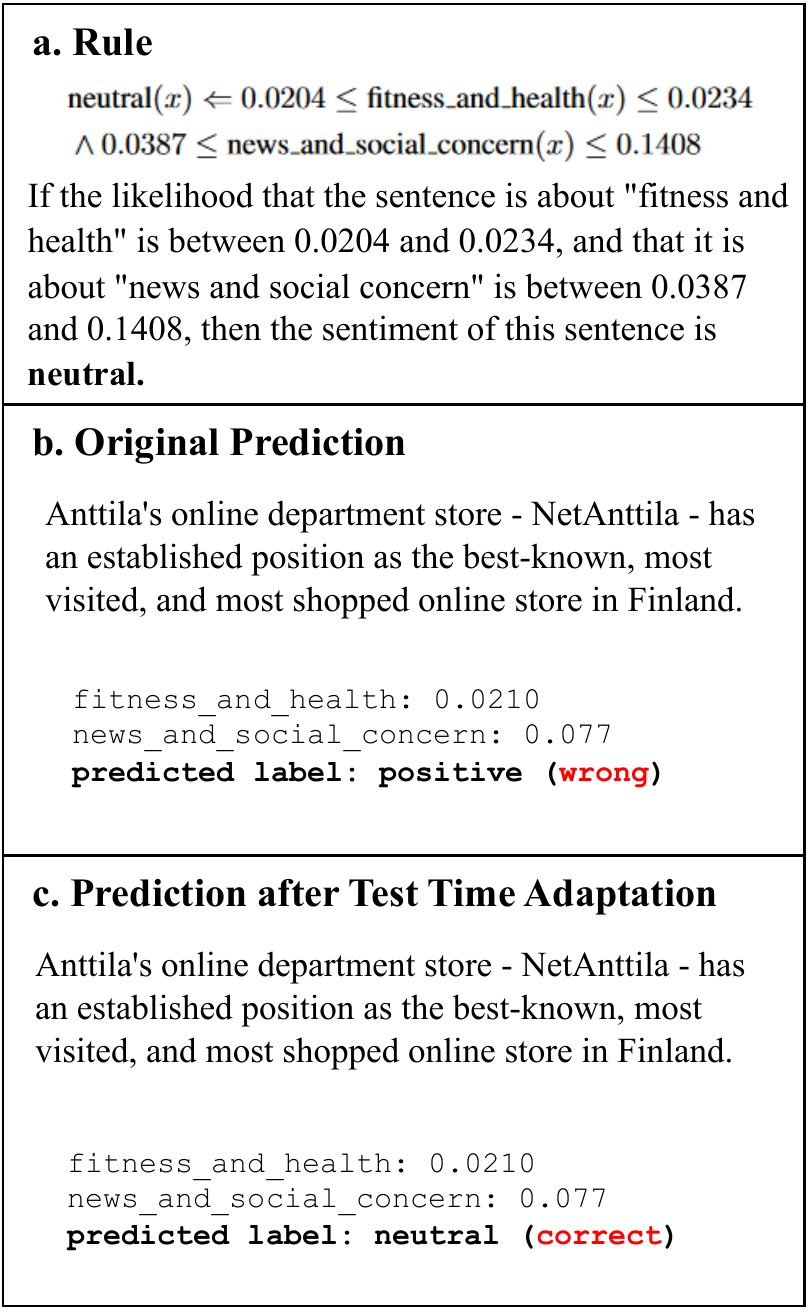}
         \caption{Sentiment analysis.}
         \label{fig:qual:sentiment}
     \end{subfigure}
     \hfill
     \begin{subfigure}[b]{0.32\textwidth}
         \centering
         \includegraphics[width=\textwidth, height=1.52\textwidth]{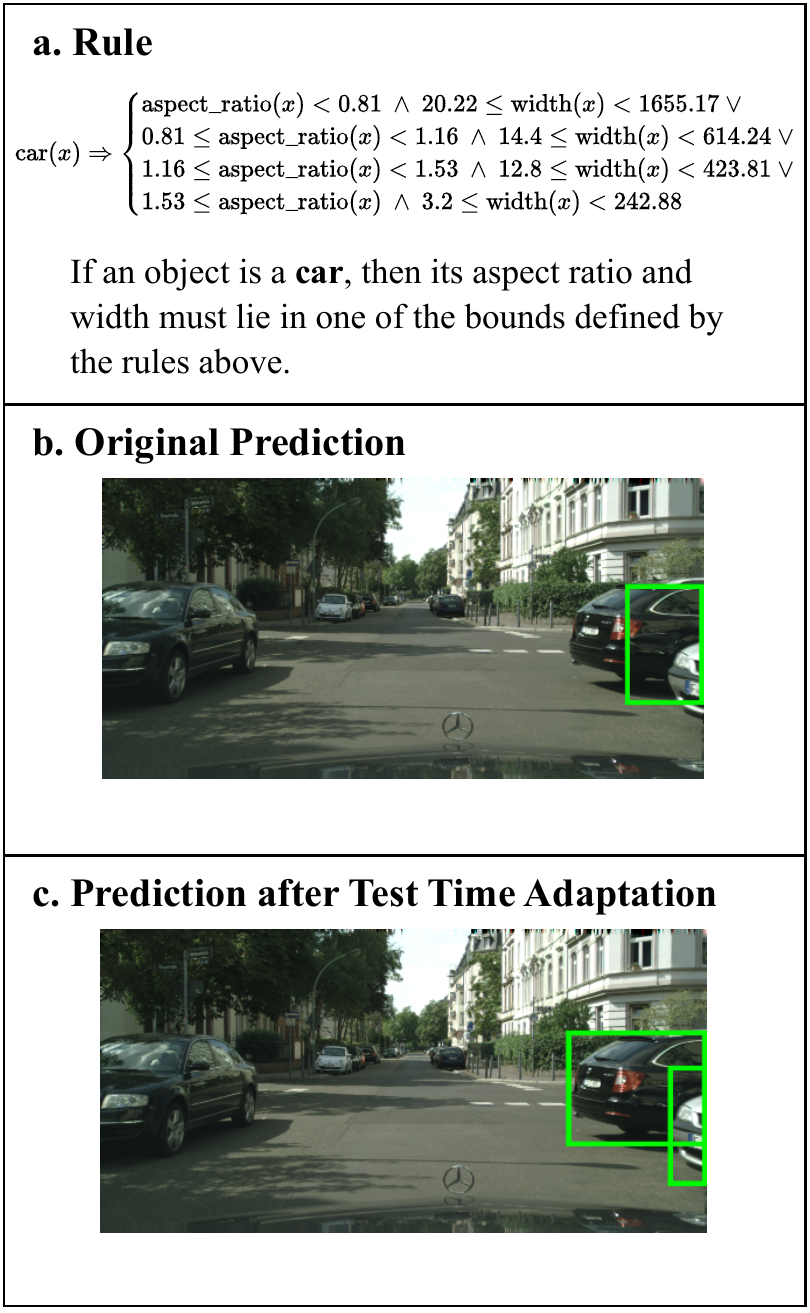}
         \caption{Object detection.}
         \label{fig:qual:obj}
     \end{subfigure}
        \caption{Qualitative results on time series imputation, sentiment analysis, and object detection tasks (those for the tabular and image classification tasks are in Appendix \ref{sec: addition_quantitative}). For each task, we show (a) a rule generated by \tool{}, (b) a model's prediction violating the rule, and (c) the model's prediction satisfying the rule after test-time adaptation.
        Each of these rules satisfies our three desiderata: they are valid since by construction they hold for 98\% of the data; they are expressive enough to find faults in models; and they are generated at scale without human supervision on individual rules.}
        \label{fig:qual}
        \vspace{-0.1in}
\end{figure*}

We evaluate \tool{} on datasets and models for five different tasks.
We present metrics and qualitative aspects of statistical quantile rules generated by the framework from various datasets.
We measure how well existing models trained on those datasets perform with respect to the generated rules.
We also evaluate the effectiveness of adapting the models at test time to reduce rule violations.
We include some additional discussions on the uses and concerns of the generated rules in Appendix~\ref{app:discussion}.

{\bf Benchmark Tasks and Models.}
We consider five different tasks: tabular classification, image classification, object detection, time-series imputation, and finally semantic analysis.
We describe the datasets below and further detail them in Appendix~\ref{app:domains}.

For tabular classification, we consider the data from the Cardiovascular Disease dataset \citep{cardio}
over which we train the FT-Transformer model \cite{gorishniy2021revisiting}.

For image classification, we use a ResNet-34 model trained over the original ImageNet dataset \cite{deng2009imagenet} but use ImageNet-X \cite{idrissi2022imagenet} for generating the rules and evaluating violations.
The ResNet-34 model is trained to classify between the 17 metaclasses defined by ImageNet-X.

For object detection, we consider the object detector component of the EfficientPS model \cite{mohan2021efficientps}. Specifically, we leverage a version of the EfficientPS model which is pretrained over the KITTI self-driving dataset \cite{geiger2013vision}\footnote{The pretrained model is downloaded from \url{https://github.com/DeepSceneSeg/EfficientPS}}.
We evaluate the model over the validation splits of the KITTI, \cityscape{} \cite{cordts2016cityscapes}, and \cityscape{}$_\text{rain}$ \cite{tremblay2020rain}, a version of \cityscape{} augmented with heavy rains.

For time series imputation, we trained one state-of-the-art time series imputation model, SAITS \citep{du2023saits} on the Physionet Challenge 2012 dataset \citep{silva2012predicting} (\physionet{} for short). After using the pre-processing scripts from \citep{tashiro2021csdi}, the resulting time series samples contain around 80\% missing entries. The goal of this task is to impute those missing values.

For semantic analysis, we use the pretrained FinBERT model \citep{araci2019finbert} on the Financial PhraseBank dataset \citep{malo2014good} (\PhraseBank\ for short). The goal of this task is to predict the sentiment (negative, neutral, or positive) of each statement about financial news in this dataset.

\subsection{Evaluating Models using Statistical Quantile Rules}
\label{sec:eval:rq1}
To evaluate how well machine learning models capture statistical quantile rules, we synthesize a suite of rules over their corresponding training data and measure the total number of violations of these rules.
We describe the setup for each task below and summarize the results in Table \ref{tab:res:rq1}.

{\bf Tabular Classification.}
The type of rules considered varies across different tasks and is specified via the rule schema.
For the tabular classification task, we consider a two-sided form of \textit{logic quantile rules} with the statistic $\phi$ as the F1 score of the rule over a minibatch.
As discussed in Section~\ref{sec:rules:classes}, these rules are expressed as boolean formulae over the features from the respective datasets.

We use the columns in the cardiovascular dataset as the features over which the formulae can be generated.
While columns such as `${\sf smoke}$' are boolean-valued, others such as `${\sf age}$' are not.
We bucket the values of such columns into 8 buckets to produce 8 boolean-valued features for values belonging in each bucket.
This results in a total of 52 features over which the rules are generated.
We generate 300K rules, of which 400 are selected (200 rules per class) for evaluation in the manner described in Algorithm \ref{alg:compute_stat}.
We show an example of a rule and its violating minibatch in Figure \ref{fig:qual:tab} in Appendix \ref{sec: addition_quantitative}.

{\bf Image Classification.}
Similar to the previous task, we consider two-sided \textit{logic quantile rules} with the F1 score as the summarization statistic for each minibatch.
The features are obtained from the boolean-valued annotations in ImageNet-X, such as ${\sf background}$ or ${\sf pose}$.
This yields 16 features over which rules are generated.
We show an example of a rule and its violation in Figure~\ref{fig:qual:img} in Appendix~\ref{sec: addition_quantitative}.

\begin{table*}[!t]
\centering
\footnotesize
\begin{tabular}{c c c c c c} \toprule
    \textbf{Task} & \textbf{\makecell{Tabular  \\ Classification}} & \textbf{\makecell{Image \\Classification}} & \multicolumn{1}{c}{\textbf{Object Detection}} & \textbf{\makecell{Time Series\\ Imputation}} & \textbf{\makecell{Sentiment\\ Analysis}} \\ \midrule
    Model & FT-Transformer & ResNet-34 & EfficientPS & SAITS & FinBERT \\ \hline
    Rule Class & Logic & Logic & Neural & Neural & Logic \\ \hline
    Sample Size & 4096 & 256 & 1 & 1 & 128 \\ \hline
    Dataset {\bf R} & Cardiovascular$_{\text{train}}$ & ImageNet$_{\text{rule-train}}$ & KITTI$_{\text{train}}$ & \physionet{}$_{\text{train}}$ & \PhraseBank{}$_{\text{train}}$ \\ \hline
    Dataset {\bf V} & Cardiovascular$_{\text{valid}}$ & ImageNet$_{\text{rule-valid}}$ & - & \physionet{}$_{\text{valid}}$ & \PhraseBank{}$_{\text{valid}}$ \\ \hline
    Dataset {\bf T} & Cardiovascular$_{\text{test}}$ & ImageNet$_{\text{valid}}$ & Cityscapes$_{\text{rainy,valid}}$& \physionet{}$_{\text{test}}$ & \PhraseBank{}$_{\text{test}}$ \\ \hline
    \# Total Rules & 292,129 & 73,032 & 252 & 35 & 7878 \\ \hline
    \# Selected Rules & 400 & 340 & 252 & 35 & 158 \\ \hline
    \# Total Violations & 26083$\pm$56 & 7716$\pm$318 & 8,108 & 157772$\pm$18697.80 & 578 \\ \hline
     \makecell{\# Rules Violated \\ per Sample} & 389.29$\pm$0.83 & 42.62$\pm$1.76 & 16.21 &197.46$\pm$23.40 & 0.59 \\
    \bottomrule 
\end{tabular}
\caption{Results of generating rules and evaluating models against them.
Rules are generated over each training dataset using the specified sample size.
For sample sizes larger than one, we randomly sample a batch from the dataset.
The validation datasets are used to sample a subset of generated rules for evaluation.
We count the number of selected rules violated by the model on each test sample.
The last two rows show the sum and average of these counts over all test samples respectively.
}
 \label{tab:res:rq1}

\end{table*}

Since we evaluate rule violations over the ImageNet validation set, we split the training set of the ImageNet-X dataset into a \textit{rule-training} set and a \textit{rule-validation} set using an 80\%/20\% split.
For each meta-class, we also select 20 rules (i.e., 340 rules in total out of around 73K) in accordance with Algorithm~\ref{alg:compute_stat}.

{\bf Object Detection.}
Due to the nature of the predictions in this task, instead of logic quantile rules, we consider a two-sided form of \textit{neural quantile rules} for a total of 6 statistics related to the bounding box predictions made by the model.
These statistics include the aspect ratio, width, height, area, the x-coordinate of the center, and the y-coordinate of the bottom of the bounding box.
Each statistic is defined with respect to individual bounding boxes.
Statistics are also considered in pairs. For each class, we consider all pairs of statistics $(s_1, s_2)$. For each pair, we group objects of that class by $s_1$ into four buckets. Within each bucket, we learn the 98 percentile bounds for statistic $s_2$.
This results in generating rules as seen in Figure \ref{fig:qual:obj}(a), where $s_1$ is the aspect ratio and $s_2$ is the width of each bounding box.
A violation of this rule is shown in Figure \ref{fig:qual:obj}(b).
We learn 252 rules using these 6 statistics for 7 classes that the object detector is trained to predict.
In all cases, the 98 percentile bounds for the rules are learned over the KITTI training dataset.

{\bf Time Series Imputation.}
The imputation model outputs estimates of the missing entries.
Therefore, for this task, we consider the set of statistics to be the features in the dataset.
For each feature, we consider a two-sided form of \textit{neural quantile rules} over the imputed values.
As mentioned in Appendix \ref{app:domains}, there are 35 features in the \physionet{} dataset resulting in 35 generated logic rules.
For each rule, the 98 percentile bounds of the statistics are learned over the non-missing values of the training split of the \physionet{} dataset.
Figure \ref{fig:qual:time_series}(a) presents an example of a generated rule.
This rule shows the bounds learned for the ``Glasgow Coma Score'' feature, which also matches the universally accepted range of this feature in the medical domain \citep{teasdale1974assessment}.
We show an example of an imputed value violating this rule in Figure \ref{fig:qual:time_series}(b) where it exceeds the upper bound.

{\bf Semantic Analysis.}
Similar to the previous two tasks, two-sided {\em neural quantile rules} are considered over each sentence in the \PhraseBank{} dataset.
For each sentence, we use pretrained models to extract features that make up the statistics for rules to be generated.
We use the pretrained DistilRoBERTa model \citep{hartmann2022emotionenglish} to calculate the likelihood of each of the 7 basic emotions, and the Roberta model pretrained on 11,267 tweets \citep{dimosthenis-etal-2022-twitter} to calculate the likelihood of each of the 19 Twitter topics for each sentence.
This results in 26 features generated for each sentence.
Similar to the object detection task, for each class we generate rules over each feature, as well as over each pair of features, producing 7.8K rules in all.
We then select 158 rules in accordance with Algorithm \ref{alg:compute_stat}.
Figures \ref{fig:qual:sentiment}(a) and \ref{fig:qual:sentiment}(b) show an example of a rule and its violation respectively.

\begin{table*}
    \centering
    \footnotesize
    \begin{tabular}{@{}c c @{\ \ \ } c  c @{\ \ \ } c  c c @{\ \ \ } c  c @{\ \ \ } c c @{\ \ \ } c  c @{}}\toprule 
  {\bf Task} & \multicolumn{2}{c}{{\bf \makecell{Tabular \\ Classification}}} & \multicolumn{2}{c}{{\bf \makecell{Image \\ Classification}}} & \multicolumn{2}{c}{{\bf \makecell{Object \\ Detection}}} & \multicolumn{2}{c}{{\bf \makecell{Time series \\ Imputation}}} & \multicolumn{2}{c}{{\bf \makecell{Sentiment\\ Analysis}}} \\ \midrule
    Dataset  & \multicolumn{2}{c}{\healthdata} & \multicolumn{2}{c}{\imagenetx} & \multicolumn{2}{c}{\cityscape-rainy}& \multicolumn{2}{c}{\physionet{}}& \multicolumn{2}{c}{\PhraseBank{}} \\
    Metric   & AUC & \% Reduced & Accuracy & \% Reduced     & mAP   & \% Reduced & \makecell{MSE \\($\times$10$^{-3}$)} &\% Reduced & Accuracy & \% Reduced \\ \midrule
    No adapt & 80.10 & -              & 80.47 &  -               & 25.49 & - &       7.7 &   - & 84.02 & -  \\ \midrule
    BN       & 80.10 & 0.0\%         & 80.47 &  2.9$\pm$3.4\% & 25.49 & -0.0$\pm$0.0\%& 7.7 & 0.0$\pm$0.0\%  &84.02 & 0.0$\pm$0.0\% \\
    \tent    & 80.10 & -0.1$\pm$0.0\% & {\bf 80.51} &  0.7$\pm$5.3\%  & 25.52 & -1.4$\pm$0.8\% & 8.0 &  -0.2$\pm$0.9\% & {\bf 86.08}&14.7$\pm$0.0\% \\
    RPL      & 80.09 & -0.1$\pm$0.0\% & 80.48 & -0.8$\pm$6.5\% & 25.70 &  0.2$\pm$0.3\% & 7.4& 18.7$\pm$16.2\% & 84.85 & 7.3$\pm$0.2\% \\
    CPL      & 80.10 & -0.1$\pm$0.0\% & 80.44 & -0.1$\pm$6.0\% & 24.95 & -0.1$\pm$0.1\% & 5.6& 12.7$\pm$6.5\% & 85.77 & {\bf 15.7$\pm$0.1\%}\\
    \memo    & 80.10 & 1.8$\pm$0.0\%  & 80.47 & 11.6$\pm$4.1\%  &  -     & -  & - & - & {\bf 86.08} &  15.0$\pm$0.2\%  \\ \midrule
    Ours     & {\bf 80.12} & {\bf 38.1$\pm$0.0\%} & {\bf 80.51} & {\bf 22.2$\pm$4.5\%}  & {\bf 27.01} & {\bf 31.4$\pm$0.2\%} & {\bf 5.2} & {\bf 68.7$\pm$8.1\%} & {\bf 86.08} & {\bf 15.7$\pm$0.1\%} \\ \bottomrule
    \end{tabular}
    \caption{Results of reducing rule violations using different test-time adaptation methods for our three tasks.}
    \label{tab:rq2}
\end{table*}

\textbf{Results.}
Table \ref{tab:res:rq1} shows the total and per sample number of rule violations.
For the time series imputation task, the number of rule violations is large (up to around 158K) despite the smaller number of rules (35).
This indicates that the state-of-the-art time series imputation models are far from perfect.
Apart from this, the image and tabular classification tasks also violate a larger number of rules per sample.
Note that the rules for the tabular and image classification tasks are primarily a measure of the consistency of a batch of predictions with respect to the training data.
Since we have 200 rules per class for tabular classification to measure this consistency as opposed to 20 for image classification, misclassifications can contribute to a larger number of rule violations in the former task.

On the other hand, there are fewer violations per sample in the object detection and sentiment analysis tasks.
Only a minority of the generated rules are violated by the predictions of the object detector. We show the violation results on the \cityscape{}-rainy dataset in Table \ref{tab:res:rq1}, and results on \kitti\ and \cityscape{} in Appendix \ref{sec: addition_quantitative}.

\subsection{Adapting Models using Statistical Quantile Rules}
\label{sec:eval:rq2}
We next consider reducing the number of test-time violations of the rules shown in Table \ref{tab:res:rq1}.
We do so by using the rule-driven test-time adaptation technique presented in Section \ref{sec:framework}.
We compare \tool{} with state-of-the-art test-time adaptation techniques as baselines.

{\bf Baselines.}
Most test-time adaptation methods, surveyed in Section \ref{sec:related_work}, are orthogonal to our work.
We therefore only compare against the following most relevant baselines:
\begin{itemize}[leftmargin=*,nolistsep, nosep]
    \item \textbf{Batch normalization}: Batch normalization (BN) \cite{ioffe2015batch} computes statistics of the batch normalization layers during test time rather than reusing the learned statistics during the training phase. 
    \item \textbf{Entropy minimization-based methods}: \tent\ \cite{wang2020tent} minimizes the entropy of the model predictions, but only focuses on the classification problem. We, therefore, follow the analysis from \cite{goyaltest} to minimize the negative L2 norm of the model output for the regression problem in the object detection task.
    \item \textbf{Pseudo-label-based methods}: \rpl\ (RPL) \cite{rusak2021if} and \cpl\ (CPL) \cite{goyaltest} generate pseudo labels as supervisions for test-time adaptations. 
    \item \textbf{Data augmentation-based methods}: \memo\ \cite{zhang2021memo} generates multiple augmented versions of a single test sample and minimizes the entropy of the model output across these samples. Since it primarily deals with classification models, generalizing it to object detection models and time series imputation models requires non-trivial efforts. We, therefore, ignore the comparison between \memo\ and \tool{} in the object detection and time series imputation tasks. 
\end{itemize}

For the baseline methods, we follow the default setups that perform test-time adaptation for a few epochs since overfitting can occur with more epochs.
We use up to 60 epochs for \tool{}.
We follow the default setups of test-time adaptation by only fine-tuning the statistics of the batch normalization layers rather than the entire model.
We also perform an ablation study in Appendix \ref{sec: ablation} to analyze the effect of fine-tuning the entire model with \tool{}.

{\bf Results.}
The evaluation results are shown in Table \ref{tab:rq2}.
We report two metrics for all methods: the model performance and the percentage of rule violations reduced (``\% Reduced'') with respect to the total violations reported in Table \ref{tab:res:rq1} after adaptation.
For the model performance metrics, we report the AUC score for tabular classification (since it is a binary classification task), the prediction accuracy for image classification, the Mean Average Precision (mAP) for object detection, the Mean Square Error (MSE) for time series imputation and the accuracy for sentiment analysis.
\tool{} is able to significantly reduce the number of rule violations at test-time (by up to 68.7\% in the \physionet{} dataset), which can also lead to up to 32\% relative performance improvement in the model in the \physionet{} dataset (reducing the MSE from 7.7$\times$10$^{-3}$ to 5.2$\times$10$^{-3}$) and a slight performance improvement in other datasets.
We show examples of time series imputation, sentiment analysis, and object detection in Figure \ref{fig:qual}, wherein each subfigure, the incorrect predictions violating the rule shown in (a) are depicted in (b) which are corrected in (c) after test-time adaptation.
We include more examples from the tabular and image classification tasks in Figure \ref{fig:addition_qual} in Appendix \ref{sec: addition_quantitative}.
For example, as Figure \ref{fig:qual:time_series}(c) shows, \tool{} could pull the imputed values at the 1$_{st}$ hour (the black dot) closer to the ground truth (the blue dot) after test-time adaptation. 
In contrast, the baselines negligibly affect the violations and produce models which perform slightly worse than \tool{}.

\section{Related Work}
\label{sec:related_work}

{\bf Test-time adaptation.}
Test-time adaptation aims to adapt models to new data distributions in the presence of distribution shift, which typically assumes no access to the source data or the ground-truth labels of data on the target distribution. These approaches involve minimizing the entropy of the model output \cite{wang2020tent, goyaltest}, producing pseudo-labels to perform supervised training \cite{goyaltest, rusak2021if}, or self-supervision \cite{zhangdivide, chen2022contrastive} at test-time; enforcing model output to be consistent across different augmentations of one test sample \cite{zhang2021memo}; constructing generative models for producing labeled test samples \cite{li2020model}; etc. To our knowledge, none of the existing test-time adaptation methods take into account enforcing the model output to align with statistical quantile rules. In addition, most of these techniques leverage strategies orthogonal to ours, such as meta-learning \cite{xiao2021learning} or generative models \cite{li2020model}. 
Furthermore, most of them are limited to classification tasks while our method is applicable to generic tasks, including both classification and regression tasks. 

{\bf Injecting rules into ML models.} There have been many efforts to effectively inject rules or domain knowledge into neural nets. One widely adopted strategy is to add a term encoding violations of logic rules to regularize the objective function. For instance, \cite{hu2016harnessing} designed a teacher-student framework for jointly learning from labeled samples and logic rules,  \cite{seo2021controlling} jointly learns embeddings of logic rules and training data, and \cite{ganchev2010posterior} regularizes the model posteriors with constraints on data. Some other solutions in this area include penalizing bias rules through adversarial learning \cite{zhang2018mitigating} and solving a constrained optimization problem during training by regarding logic rules as constraints \cite{fioretto2021lagrangian, narasimhan2018learning}. However, all of these approaches demand manually specified rules as input, whereas our methods can automatically learn symbolic rules and their statistical bounds. Moreover, the state-of-the-art solutions to integrate rules into ML models primarily focus on enhancing supervised learning whereas our method is able to deal with a more challenging setting, i.e., test-time adaptation. As mentioned above, there is no supervision from data during test-time adaptation and thus it is impossible to regularize model training with rules.

{\bf Rule learning.}
The problem of synthesizing first-order logic rules from data has been extensively studied in the literature on inductive logic programming (ILP).
Techniques such as ILASP \cite{law2020ilasp} and Prosynth \cite{popl20} leverage pure symbolic reasoning to search logic rules that can produce expected answers in a given database. Works such as NeuralLP \cite{yang2017differentiable} and NLIL \cite{yang2019learn} attempt to leverage neural networks to guide the search for feasible logic rules. NLIL can also learn one simple statistic, the confidence score, for each synthesized rule. However, it is less expressive than the arbitrary statistics captured by our framework. Besides, confidence scores are calculated over the entire dataset rather than over mini-batches of data. We therefore cannot check their consistency between different portions of the data. 

{\bf Weakly supervised learning with logic rules.} Similar to test-time adaptation, weakly supervised learning is also applicable when no ground-truth labels exist. One weak supervision strategy is to employ logic rules to annotate unlabeled samples, which is then followed by regular supervised learning \cite{ratner2016data, ratner2017snorkel}. However, to our knowledge, state-of-the-art rule-based weakly supervised learning approaches can only handle classification tasks. In contrast, our solutions are applicable in very general machine learning settings, including both classification and regression tasks. In addition, although some works such as \cite{varma2018snuba} can automatically derive logic rules as weak supervision signals, they are limited to reasoning about symbolic rules. Hence, they are not applicable to our setting where statistics are necessary.

\section{Conclusion and Future Work}

We formalized statistical quantile rules as a means of characterizing basic errors inconsistent with training data and defined the problem of extracting such rules at scale.
We proposed \tool{}, a general framework to generate a large number of such rules for a given dataset and evaluate violations of these rules by a model.
Through our extensive empirical studies, we found that machine learning models do not always learn statistical rules inferred from data but can be adapted to correct these rule violations by leveraging rule-based test-time adaptation.
In the future, we intend to evaluate statistical quantile rules more widely over modern models.
We also intend to explore more uses of these rules, 
like more effective ways to train models to reduce rule violations and
using them for unsupervised learning.

\section*{Acknowledgements}
We thank the reviewers for their feedback.
This research was supported in part by NSF grants \#1836936 and \#2107429.

\bibliography{references}
\bibliographystyle{icml2023}

\newpage
\appendix
\onecolumn
\section{Additional Details of Jaccard Index}\label{sec: app_tech}
As introduced in Section \ref{sec: rule_learning}, the Jaccard index is used for selecting consistent statistical rules between the training and validation sets. To illustrate how to compute the Jaccard index, we revisit the running example in Section \ref{sec: rule_learning} which defines the following schema:
\begin{align*}
    Y & \in \{ \sf{car},\ \sf{person},\ \sf{rider},\ \ldots \}, \\
    \phi & \in \{ \sf{aspect\_ratio},\ \ldots \},\ \ \ \gamma = 0.98, \\
    \yvar &= Y \Rightarrow \theta_{\sf lb} \leq \phi(\xvar) \leq \theta_{\sf ub},
\end{align*}

which can be evaluated on both the training set and validation set, leading to the following instantiations of the two-sided quantile rules on the training set and validation set respectively:
\begin{align*}
    \yvar &= Y \Rightarrow \theta_{\sf lb, train} \leq \phi_{\text{train}}(\xvar) \leq \theta_{\sf ub, train}\text{, for training set}, \\
    \yvar &= Y \Rightarrow \theta_{\sf lb, valid} \leq \phi_{\text{valid}}(\xvar) \leq \theta_{\sf ub, valid}\text{, for validation set}.
\end{align*}

The Jacaard index between the above two rules is thus formulated as follows:
\begin{align*}
    \text{Jaccard}(\phi_{\text{train}}, \phi_{\text{valid}}) & = \frac{[\theta_{\sf lb, train}, \theta_{\sf ub, train}] \bigcap [\theta_{\sf lb, valid}, \theta_{\sf ub, valid}] }{ \left[ \theta_{\sf lb, train}, \theta_{\sf ub, train} \right] \bigcup [\theta_{\sf lb, valid}, \theta_{\sf ub, valid}]} \\
    & = \frac{\min\{\theta_{\sf ub, train}, \theta_{\sf ub, valid}\} - \max\{\theta_{\sf lb, train}, \theta_{\sf lb, valid}\}}{\max\{\theta_{\sf ub, train}, \theta_{\sf ub, valid}\} - \min\{\theta_{\sf lb, train}, \theta_{\sf lb, valid}\}}.
\end{align*}

For one-sided quantile rules, i.e., either $\theta_{\sf lb}=-\infty$ or $\theta_{\sf ub}=\infty$, then we take the lower bound or upper bound of the statistics $\phi(\xvar)$, which is calculated from the entire dataset to replace $\theta_{\sf lb}$ or $\theta_{\sf ub}$.

\section{Test-time Adaptation Loss Functions}\label{sec: two-side-loss}

Recall the intuition behind the loss function mentioned in Equation~\ref{eq:loss_general}:
The loss $l_W(r, X)$ for rule $r$ of a model with parameters $W$ and its output $X$ is zero if $X$ satisfies $r$, and non-zero if it doesn't.
The non-zero value assigned depends on the class of the quantile rule.
Equation~\ref{eq:loss_general} shows the loss for the basic form of quantile rules.
We now define the loss functions used for test-time adaptation for some extensions of quantile rules used in this paper.

\begin{enumerate}
    \item \textbf{Two-sided Quantile Rules.}
    We extend the definition from \eqref{eq:loss_general} for two-sided quantile rules.
    Assume a two-sided quantile rule $r := \theta_{\sf lb} \leq \phi(X) \leq \theta_{\sf ub}$.
    We define the loss for such a rule as so:
    \begin{align}
        \label{eq:loss_general_two_sided}
        l(r; X) = \begin{cases}
        0 & X \text{satisfies}\ r \\
        \min \{ (\theta_{\sf lb} - \phi(X))(\theta_{\sf ub} - \phi(X)), 1 \} & \text{otherwise} \\
        \end{cases}
    \end{align}
    Note that the non-zero loss is a quadratic function of $\phi(X)$ with the upper and lower bounds being the roots, but with the output of $\phi(X)$ clipped at 1.
    This ensures a positive non-zero loss for a rule violation when the value of $\phi(X)$ is either less than $\theta_{\sf lb}$ or more than $\theta_{\sf ub}$.

    \item \textbf{Logic Quantile Rules.}
    Assume a two-sided logic quantile rule $r := \theta_{\sf lb} \leq \phi(\psi(X)) \leq \theta_{\sf ub}$, where $X$ is the output of a model with parameters $W$ over a minibatch of samples.
    The loss for such a rule is dependent on the statistic $\phi$ since it is a statistic that aggregates the result of $\psi(X)$.
    For our experiments, we use the F1 score as this statistic.
    However, as discussed in Section~\ref{sec:uses}, the F1 score is non-differentiable, so we instead use a surrogate F1 loss defined in \cite{benedict2021sigmoidf1}.
    Let this loss function be termed $L_{\sf F1}$.
    We then define the loss for logic quantile rules as so:
    \begin{align}
        \label{eq:loss_general_two_sided}
        l(r; X) = \begin{cases}
        0 & X \text{satisfies}\ r \\
        \min \{ (\theta_{\sf lb} - L_{\sf F1}(\psi(X)))(\theta_{\sf ub} - L_{\sf F1}(\psi(X))), 1 \} & \text{otherwise} \\
        \end{cases}
    \end{align}
\end{enumerate}

\section{Supplementary Experiments}
\subsection{Ablation study}\label{sec: ablation}
Note that in the experiments, during the rule-based test time adaptation phase, we only fine-tune the batch normalization layers of the models. The comparison between fine-tuning the whole model and fine-tuning batch-normalization layers has been studied for many state-of-the-art test time adaptation methods, such as \cite{wang2020tent}. We therefore also conduct this experiment as our ablation study for the Tabular classification task, Image classification task, and object detection task on the \cityscape-rainy dataset.

The results are presented in Table \ref{tab: ablation}. This table clearly shows that it is not ideal for fine-tuning the whole model during the rule-based test time adaptation phase since it can significantly drop the model performance by up to 9\%. Therefore, same as prior test time adaptation methods, it is reasonable to fine-tune batch normalization layers rather than the entire model for rule-based test time adaptation.

\begin{center}
\begin{table*}[t]
    \centering
    \small
    \begin{tabular}{ c | c | c | c | c | c | c}\toprule
    \multirow{2}{*}{} & \multicolumn{2}{c|}{Tabular classification} & \multicolumn{2}{c|}{Image classification}& \multicolumn{2}{c}{\makecell{object detection \\ (\cityscape-rainy)}} \\ 
    \cline{2-3}\cline{4-5}\cline{6-7} %
    & AUC score  & \% Reduced  & Accuracy  & \% Reduced & mAP  & \% Reduced \\ \hline
     Our method (fine-tuning whole model) &71.17  & \makecell{30.61$\pm$3.5\%} & 80.40 &23.28$\pm$1.40 & 26.87& 31.3$\pm$0.5\%\\
     Our method &80.12  & \makecell{38.1$\pm$0.0\%} &80.51  &\makecell{22.17$\pm$4.5\%}& 27.01& 31.4$\pm$0.2\%\\  \bottomrule
    \end{tabular}
    \caption{Results for fine-tuning the whole models VS fine-tuning batch normalization layers}
    \label{tab: ablation}
\end{table*}
\end{center}

\subsection{Additional quantitative results}\label{sec: addition_quantitative}
In Table \ref{tab:res:add_rq1}, we present the results of generating and evaluating rules on the \kitti\ and \cityscape\ datasets respectively. 
By comparing the results of the \cityscape-rainy dataset against the above results, we can observe that the violations increase over datasets of a different distribution like \cityscape{}.
This thus implies that the consistency of the model's predictions to the training data worsens over datasets that may be out of distribution. In Table \ref{tab:add_rq2}, we show the results after performing test-time adaptations over the \kitti\ and \cityscape\ datasets, which still suggests that our methods can reduce more rule violations than other methods without hurting the model performance.

\begin{table*}
\centering
\footnotesize
\begin{tabular}{c c c} \toprule
    \textbf{Task} & 
    \multicolumn{2}{c}{\textbf{Object Detection}} 
    \\ \midrule
    Model &
    EfficientPS & EfficientPS 
    \\ \hline
    Rule Class & 
    Neural & Neural  
    \\ \hline
    Sample Size & 
    1 & 1 
    \\ \hline
    Dataset {\bf R} & 
    KITTI$_{\text{train}}$ & KITTI$_{\text{train}}$ 
    \\ \hline
    Dataset {\bf V} 
    & - & - 
    \\ \hline
    Dataset {\bf T} 
    & KITTI$_{\text{valid}}$ & Cityscapes$_{\text{valid}}$ 
    \\ \hline
    \# Total Rules 
    & 252 & 252 
    \\ \hline
    \# Selected Rules & 
    252 & 252 
    \\ \hline
    \# Total Violations & 
    1,561 & 7,673 
    \\ \hline
     \makecell{\# Rules Violated \\ per Sample}  
     & 7.80 & 15.34 
     \\
    \bottomrule 
\end{tabular}

\caption{Results of generating rules and evaluating models against them on \kitti\ and \cityscape\ dataset respectively.
}
 \label{tab:res:add_rq1}
 \vspace{-0.15in}
\end{table*}

\begin{table*}[t]
    \centering
    \footnotesize
    \begin{tabular}{@{}c c @{\ \ \ } c  c @{\ \ \ } c @{}}\toprule 
  {\bf Task}  
  & \multicolumn{4}{c}{{\bf Object Detection}} 
  \\ \midrule
    Dataset  & 
    \multicolumn{2}{c}{\kitti} & \multicolumn{2}{c}{\cityscape} \\
    Metric   & 
    mAP   & \% Reduced     & mAP   & \% Reduced    \\
    \hline
    No adapt 
    & 44.23 &  -             & 56.15 & -              
    \\ \hline
    BN       & 
    44.23 &  0.0$\pm$0.0\% & 56.15 &  0.0$\pm$0.0\% \\
    \tent    & 
    44.84 &  0.2$\pm$0.2\% & 56.24 &  0.0$\pm$0.0\% 
    \\
    RPL      & 
    44.26 &  0.4$\pm$0.1\% & 56.67 & -0.7$\pm$0.2\% \\
    CPL      & 
    44.62 & -3.7$\pm$0.0\% & 57.13 &  0.2$\pm$0.1\% 
    \\
    \memo    
    & -     &  -             & -     & -             
    \\ \hline
    Ours     &
    44.27 &  0.1$\pm$0.2\% & 56.91 &  6.1$\pm$1.2\% 
    \\ \bottomrule
    \end{tabular}
    \caption{Results of reducing rule violations using different test-time adaptation methods for our three tasks.}
    \label{tab:add_rq2}
    \vspace{-0.15in}
\end{table*}

\subsection{Additional qualitative results}\label{sec: addition_qualitative}
\begin{figure*}[!t]
     \centering
     \begin{subfigure}[b]{0.32\textwidth}
         \centering
         \includegraphics[width=\textwidth]{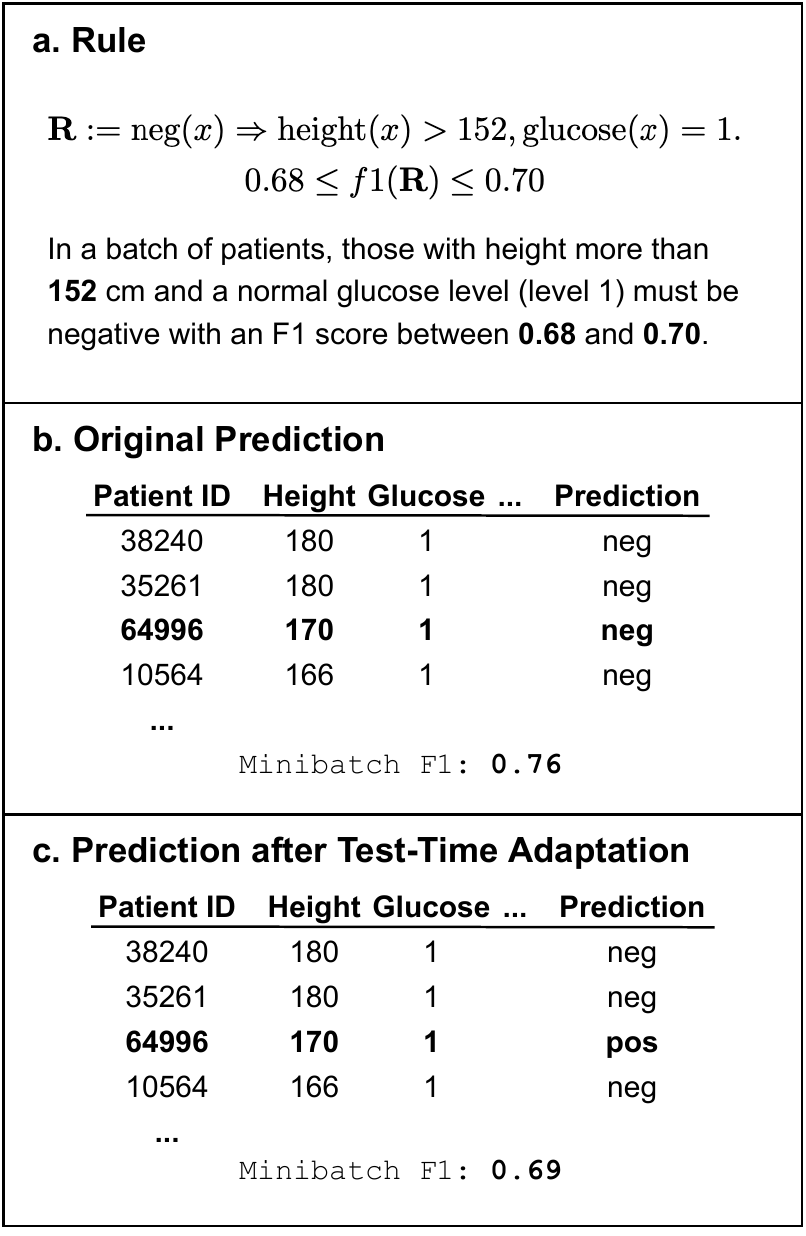}
         \caption{Tabular Classification task.}
         \label{fig:qual:tab}
     \end{subfigure}
     \begin{subfigure}[b]{0.32\textwidth}
         \centering
         \includegraphics[width=\textwidth]{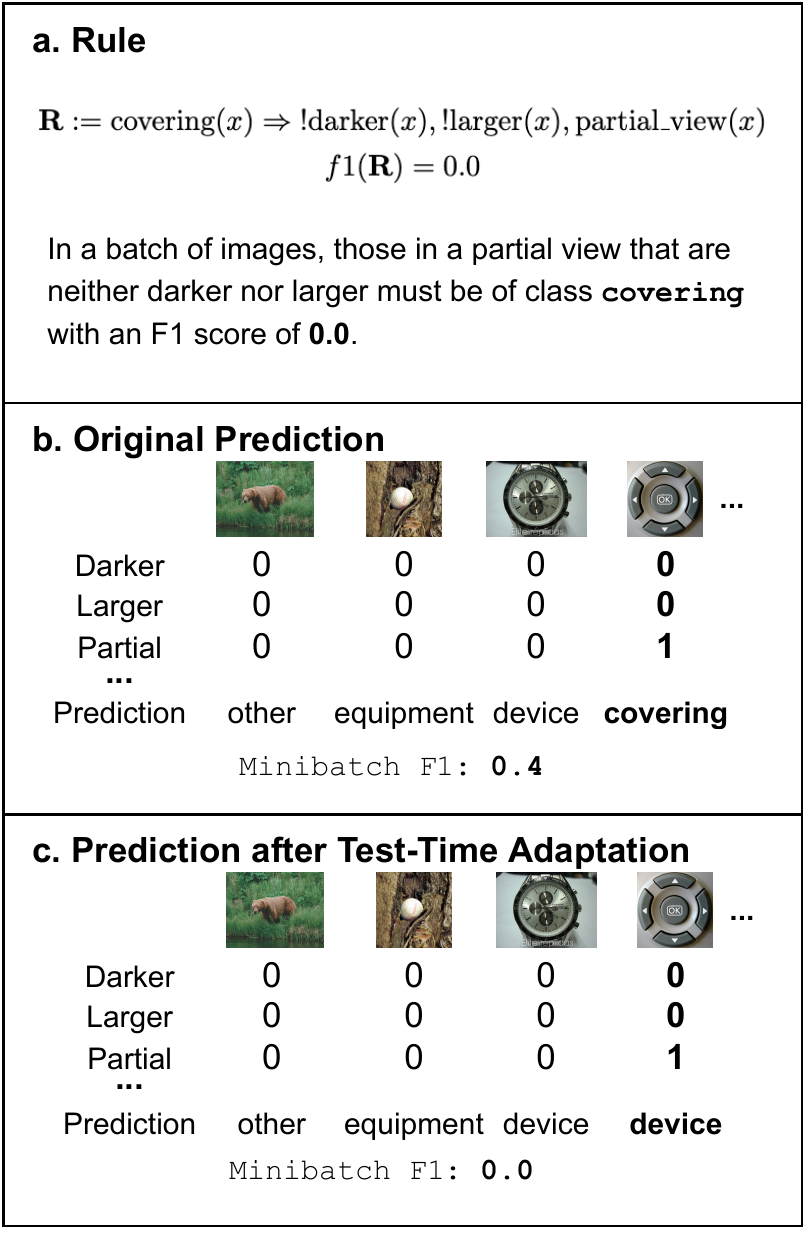}
         \caption{Image Classification task.}
         \label{fig:qual:img}
     \end{subfigure}
        \caption{Additional qualitative results on Tabular Classification and Image Classification task. For each of the three tasks, we show (a) a rule generated by \tool{}, (b) a model's prediction that violates the rule, and (c) the model's prediction satisfying the rule on the sample after test-time adaptation.
        Each of these rules satisfies our three desiderata: they are valid since by construction they hold for 98\% of the data; they are expressive enough to find faults in models; and they are generated at scale without human supervision on individual rules.}
        \label{fig:addition_qual}
        \vspace{-0.15in}
\end{figure*}

In Figure \ref{fig:addition_qual}, we show the qualitative results for Tabular Classification and Image Classification tasks.

\section{Benchmark Tasks and Models}
\label{app:domains}

We consider three different tasks: tabular classification, image classification, and object detection.

{\bf Tabular Classification.}
\label{subsec:results:tab_class}
This task concerns classification over tabular data from the Cardiovascular Disease dataset.
The dataset has information about 70k patients in the form of 11 attributes (6 numeric attributes and 5 discrete attributes) collected at medical examinations and one binary label to indicate whether the patient has cardiovascular disease.
We split the dataset by 65\%/15\%/20\% for training, validation, and testing.
We train FT-Transformer \cite{gorishniy2021revisiting}, a state-of-the-art tabular classification model, on the training set.
This model is then adapted to the test dataset during the test-time adaptation phase.
For learning and evaluating the rules, we use 67 training, 22 validation, and 6 testing samples, each of size 4096.

{\bf Image Classification.}
\label{subsec:results:img_class}
We define this task using ImageNet-X \cite{idrissi2022imagenet}, a variant of the ImageNet dataset \cite{deng2009imagenet}.
It consists of a subset of 12k training samples and all the validation samples from the ImageNet dataset.
Each image in the ImageNet-X dataset is also associated with 16 binary human annotations to indicate the characteristics of the object in the image, such as the pose of the object or the lighting of the entire image, over which we can define logic rules. 
We use a ResNet-34 model \cite{he2016deep} trained on the training data from the original ImageNet dataset to predict the 17 metaclasses defined by \citep{idrissi2022imagenet}.
We perform test-time adaptation over the validation samples of this dataset.
For learning and evaluating the rules, we use 181 training, 79 validation, and 9 testing samples, each of size 256.

{\bf Object Detection.}
This task concerns object detection in self-driving applications over the \cityscape\ \cite{cordts2016cityscapes} and KITTI datasets \cite{geiger2013vision}.
We first consider different scenarios.
First, we adapt the EfficientPS model \cite{mohan2021efficientps} pre-trained on training data from the \kitti\ dataset \cite{geiger2013vision}\footnote{The pre-trained model is downloaded from \url{https://github.com/DeepSceneSeg/EfficientPS}} to the validation samples of that dataset. 
Additionally, we evaluate our method in distribution shift settings in which the pre-trained EfficientPS model is adapted to two versions of \cityscape\ datasets \cite{cordts2016cityscapes}, one including the normal scenes and the other augmented with heavy rains \cite{tremblay2020rain}. 
Note that EfficientPS is a panoptic segmentation model which handles multiple tasks such as object detection and semantic segmentation.  We only evaluate the object detection component of this model.
We use individual samples for training and testing.
To learn the rules, we use 855 training samples.
We use 200 test samples for testing on the KITTI benchmark, and 500 each for testing on Cityscapes and Cityscapes$_\text{rainy}$.

{\bf Time series Imputation.}
By reusing the pre-process scripts from \citep{tashiro2021csdi}, we obtain 4000 time-series samples, each of which contains 35 hourly-collected features during a 48-hour stay of one patient at an ICU. We randomly partition those time-series samples into training, validation, and test set with 70-10-20 splits. To evaluate the imputation performance, we input 40\% of the non-missing observations in the test split to the imputation model, which then outputs the imputation values for the remaining 60\% of non-missing observations. We thus compare the imputed values against the 60\% ground-truth non-missing observations to evaluate the model performance.

{\bf Sentiment Analysis.} The goal of this task is to predict the sentiment of each sentence in the \PhraseBank{} dataset \citep{malo2014good}, which is composed of 3487 sentences in the training set, 387 sentences in the validation set, and 969 sentences in the test set. Note that different from the object detection task where the statistics are collected from the object detection model itself, the statistics used for the sentiment analysis task are obtained by using other pretrained models. We therefore consider neural quantile rules of the following form to guarantee a differentiable loss:
\begin{align*}
    Y & \in \{ \sf{Positive},\ \sf{Negative},\ \sf{Neutral}\}, \ \phi_1, \phi_2 \in \{ \sf{fitness\_and\_health}, \sf{news\_and\_social\_concern},\ \ldots \}, \\
    & \theta_{\sf lb} \leq \phi_1(\xvar) \leq \theta_{\sf ub} \Rightarrow \yvar = Y , \ 1-\delta = 0.98, \text{ OR }\\
    & \theta_{\sf lb,1} \leq \phi_1(\xvar) \leq \theta_{\sf ub,1}, \theta_{\sf lb,2} \leq \phi_2(\xvar) \leq \theta_{\sf ub,2} \Rightarrow \yvar = Y , \ 1-\delta = 0.98,
\end{align*}
in which, $\sf{fitness\_and\_health}$, $\sf{news\_and\_social\_concern}$ are features extracted from pretrained models.

\section{Discussions}
\label{app:discussion}

\subsection{Capturing statistical correlations.}
By constructing the statistical quantile rules, while generating them in a scalable manner, some of the generated rules may capture spurious correlations.
However, since these rules are valid over 98\% of the data, only those spurious correlations that are also valid in 98\% of the data will be captured.
Many spurious correlations in practice do not persist in such a large fraction of the data (e.g., dogs can be correlated with being outdoors but fewer than 98\% of dogs are photographed outdoors), so these correlations will not be picked up by our approach.

\subsection{Redundancy of generated rules.}
Depending on the schema, \tool{} may generate redundant rules. However, this is only the case for logic quantile rules and does not occur in schemas where logic rules are not used, like in object detection, semantic analysis, and imputation tasks.
The validated rules from the tabular classification and image classification tasks have up to 2\% of the generated rules as redundant.
Furthermore, redundant rules don’t negatively affect the test time adaptation and only slightly increase the computational costs.

\subsection{Other uses of statistical quantile rules.}
\paragraph{Providing contexts for mispredictions.}
Since statistical quantile rules can be used to evaluate models, they can also be used to identify model mispredictions at test time where the ground truth labels are not available.
We train a linear classifier using violations of rules as features for a particular sample, and whether or not that sample was mispredicted by a model as the label.
A direct interpretation of the weights of the linear classifier indicates the rules that are most correlated with model errors.
For instance, in the object-detection task, 92\% of the predictions violating the following rule are errors:
\begin{align}\label{eq: rule_indicative}
\begin{split}
    \sf{person}(\xvar) \Rightarrow & (\sf{size}(\xvar) < 4172 \land 302.6 \leq \sf{pos}_y(\xvar) < 514.1) \lor \\
    & (4172 \leq \sf{size}(\xvar) < 13680.5 \land 348.76 \leq \sf{pos}_y(\xvar) < 513.24) \lor \\
    & (13680.5 \leq \sf{size}(\xvar) < 47795.75 \land 393.44 \leq \sf{pos}_y(\xvar) < 541.90) \lor \\
    & (\sf{size}(\xvar) \geq 47795.75 \land 407.45 \leq \sf{pos}_y(\xvar) < 764.09).
\end{split}
\end{align}

\begin{figure}[!t]
    \centering
    \includegraphics[width=0.5\textwidth]{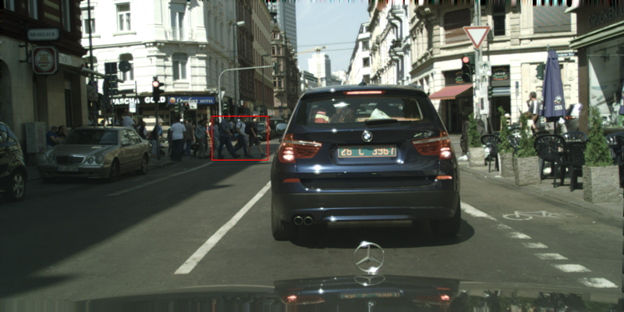}
    \caption{One example of model mispredictions violating the rule in Equation \eqref{eq: rule_indicative}}
    \label{fig:indicative_rule_example}
\end{figure}

One example of violating the above rule is shown in Figure \ref{fig:indicative_rule_example}. In this figure, the bounding box predicted to be a person violates the above rule since for the position of the bounding box, the size is too large. This thus explains the reason for this prediction being wrong.

\begin{figure*}[!t]
     \centering
     \begin{subfigure}[b]{0.32\textwidth}
         \centering
         \includegraphics[width=\textwidth]{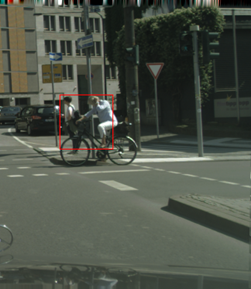}
         \caption{Predicted bounding box}
         \label{fig:metric_err1}
     \end{subfigure}
     \begin{subfigure}[b]{0.32\textwidth}
         \centering
         \includegraphics[width=\textwidth]{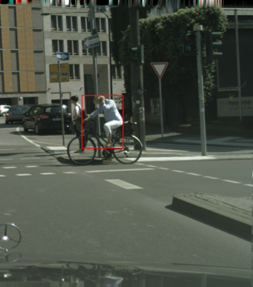}
         \caption{Ground truth bounding box}
         \label{fig:metric_err2}
     \end{subfigure}
        \caption{One example of the predicted bounding box (see Figure \eqref{fig:metric_err1}) is regarded as a correct one according to the IoU score. But this predicted bounding box is still very visually different from the ground truth bounding box (see Figure \eqref{fig:metric_err2}) and violates the rule in Equation \eqref{eq: metric_err}}
        \label{fig:metric_error}
\end{figure*}

Rules also provide contexts for mispredictions beyond what traditional metrics provide.
They can indicate the severity of errors, the frequency of certain kinds of errors over others, and similar rule violations can indicate similar kinds of errors.

\paragraph{Detecting errors that traditional metrics cannot identify.}
Even for predictions that traditional metrics deem correct, it is possible that the metric is too coarse to sufficiently evaluate that prediction. For example, the mAP metric checks the overlap (IoU score) between a predicted bounding box and a ground truth box. If it finds a ground truth box that has an overlap of more than a threshold (typically 0.5), then it checks the label of the predicted box and says the prediction is correct if the labels match. We show such an example here, where the predicted bounding box is shown in Figure \ref{fig:metric_err1}, while the ground truth is shown in Figure \ref{fig:metric_err2}. Intuitively speaking, the predicted bounding box is incorrect since it covers almost two persons in the figure as opposed to the ground truth. According to the metrics used for the self-driving model, this is a correct prediction, though it violates the following quantile rule about pedestrians since nearly all pedestrians are narrower than the bounding box would imply:
\begin{align}\label{eq: metric_err}
    \begin{split}
        \sf{person}(\xvar) & \Rightarrow (\sf{y}(\xvar) <= 433 \land 0.70 < \sf{aspect\_ratio}(\xvar) < 4.4)\\
        &\lor (433 < \sf{y}(\xvar) <= 468 \land 0.85 < \sf{aspect\_ratio}(\xvar) < 4.81) \\
        &\lor (468 < \sf{y}(\xvar) <= 513 \land 1.07 < \sf{aspect\_ratio}(\xvar) <5.0) \\
        & \lor (\sf{y}(\xvar) > 513 \land 1.20 < \sf{aspect\_ratio}(\xvar) < 5.61).       
    \end{split}
\end{align}

\end{document}